\documentclass[10pt,twocolumn,letterpaper]{article}

\usepackage[pagenumbers]{cvpr} 
\usepackage{graphicx}  
\usepackage{array}     
\usepackage{amssymb}
\usepackage{xcolor}
\usepackage{bbding}
\usepackage{pifont}
\usepackage{colortbl}
\usepackage{bbding}  
\usepackage{multirow} 

\definecolor{cvprblue}{rgb}{0.21,0.49,0.74}
\usepackage[pagebackref,breaklinks,colorlinks,allcolors=cvprblue]{hyperref}

\def\paperID{24} 
\def\confName{CVPR}
\def\confYear{2025}

\title{Cross-View Multi-Modal Segmentation @ Ego-Exo4D Challenges 2025}

\author{Yuqian Fu$^{1}$, Runze Wang$^{2}$, Yanwei Fu$^{2}$, Danda Pani Paudel$^{1}$, Luc Van Gool$^{1}$
\\$^1$INSAIT, Sofia University “St. Kliment Ohridski”, $^2$Fudan University
}

\begin{document}
\maketitle
\begin{abstract}
In this report, we present a cross-view multi-modal object segmentation approach for the object correspondence task in the Ego-Exo4D Correspondence Challenges 2025. Given object queries from one perspective (e.g., ego view), the goal is to predict the corresponding object masks in another perspective (e.g., exo view). To tackle this task, we propose a \textit{multimodal condition fusion module} that enhances object localization by leveraging both visual masks and textual descriptions as segmentation conditions. Furthermore, to address the visual domain gap between ego and exo views, we introduce a \textit{cross-view object alignment module} that enforces object-level consistency across perspectives, thereby improving the model’s robustness to viewpoint changes. Our proposed method ranked second on the leaderboard of the large-scale Ego-Exo4D object correspondence benchmark. Code will be made available at: \href{https://github.com/lovelyqian/ObjectRelator}{https://github.com/lovelyqian/ObjectRelator}.
\end{abstract}    
\section{Introduction}
\label{sec:intro}

In the process of human skill acquisition, it is crucial to be able to switch seamlessly between these two views. For example, when observing others perform a task, we could naturally imagine ourselves executing the same actions.  However, such an ego-exo or cross-view understanding has been a long-standing question for the vision community.

Recently, the introduction of Ego-Exo4D~\cite{grauman2024ego} has made significant progress in this domain by providing time-synchronized egocentric and exocentric video pairs, along with rich annotations. This opens new possibilities for exploring cross-view understanding, which holds tremendous potential for applications in areas such as virtual reality (VR), robotics, and human-computer interaction. 

In particular, we focus on the Ego-Exo4D Correspondence Challenge\footnote{\scriptsize\url{https://eval.ai/web/challenges/challenge-page/2288/}}, under the Second Joint Egocentric Vision (EgoVis) Workshop\footnote{\scriptsize\url{https://egovis.github.io/cvpr25/}} held in conjunction with CVPR 2025. This task provides synchronized ego-exo video pairs and a series of query masks indicating an object of interest in one view. The goal is to generate the corresponding object masks in the other view at each synchronized frame.

Despite being framed as a segmentation task, most of the existing segmentation models~\cite{he2017mask,cheng2022masked,lai2024lisa,rasheed2024glamm,ren2024pixellm,kirillov2023segment,li2024omg,yan2023universal,an2023temporal, wu2023object, zou2024segment, li2024onevos,guo2024x, zheng2024learning, brodermann2025cafuser, zheng2025distilling, zhou2025camsam2} cannot directly solve this problem: most of them require segmenting images according to either known object categories, text prompts, or visual cues from the target image itself. However, this task requires the model to use object information from one view as a prompt to segment the object in the other view. We notice PSALM~\cite{zhang2024psalm}, one of the multimodal segmentation models, as one exception, which enables the visual cues to be extracted from another image. While PSALM could be adapted as a baseline, it still fails to tackle the unique challenges posed by this new task: 1) \textbf{Complex backgrounds and small objects:} in the exocentric view, cluttered scenes often introduce distractors, and the target object may be extremely small—sometimes occupying only a few pixels. This increases the risk of false positives or missed detections. 2) \textbf{Significant visual shifts:} the same object can appear drastically different between the ego and exo views in terms of scale, shape, and appearance, making cross-view consistency difficult to maintain.

To address these challenges, we build our method upon the PSALM baseline, and further develop a new cross-view multimodal segmentation approach which is tailored for addressing the Ego-Exo4D correspondence task. Our method~\cite{fu2024objectrelator}, named ObjectRelator\footnote{ObjectRelator was first proposed and released in our prior work~\cite{fu2024objectrelator}.}, integrates two novel modules to solve the two unique challenges:  1) A Multimodal Condition Fusion (MCFuse) module, which leverages both visual masks and textual descriptions as segmentation conditions. This improves object localization and reduces incorrect associations. The text descriptions are not taken from the official annotations, but are instead generated by us using existing vision-language models, i.e., LLaVa~\cite{liu2023llava}. We find that the guidance from text helps to locate the correct objects. 2) A Cross-View Object Alignment (XObjAlign) module, which uses self-supervised learning to enforce object-level consistency across views, enhancing robustness to visual discrepancies. While conceptually simple, it is the first time that such object-level consistency is being explored and validated in this task.  Our approach achieves second place in IoU and first place in VA (balanced visibility accuracy) on the test set of the Ego-Exo4D Object Correspondence benchmark. Extensive quantitative and qualitative results further validate the effectiveness of our proposed approach.

\section{Approach}
\label{sec:approach}

\subsection{Task Formulation}
The Ego-Exo4D Correspondence Challenge aims to establish cross-view object-level understanding between synchronized egocentric and exocentric video pairs. Given a frame-wise object mask track (query) in one view, the task is to predict the corresponding object masks in the other view for all frames where the object is visible in both views. The dataset presents significant challenges, including long video durations (averaging 3 minutes) and small target objects occupying only a few pixels. Importantly, the task setting excludes semantic object labels, camera pose, IMU data, and depth sensors, emphasizing open-world correspondence using only RGB inputs from consumer-grade cameras. Two evaluation tracks are provided: Ego→Exo and Exo→Ego, where each track uses object masks in one view as input to predict corresponding masks in the other.

\noindent\textbf{Evaluation Metrics.} 
The Ego-Exo4D Correspondence Challenge evaluates performance using four key metrics: 1) \textbf{Intersection over Union (IoU)}, the primary metric, measures overlap between predicted and ground-truth masks across both Ego→Exo and Exo→Ego settings. 2) \textbf{Location Error (LE)} quantifies the normalized distance between the centroids of predicted and ground-truth masks. 3) \textbf{Contour Accuracy (CA)} assesses mask shape similarity after aligning their centroids. 4) \textbf{Visibility Accuracy (VA)} evaluates the model's ability to determine whether an object is visible in the target view, using balanced accuracy across all frames with query masks. Unlike the other metrics, which are computed only when the object is visible in both views, VA is calculated over all relevant frames. The final ranking is based on the average IoU across both task directions.

\subsection{Overview}
\begin{figure}[t]
    \centering
       {\includegraphics[width=1.1\linewidth]{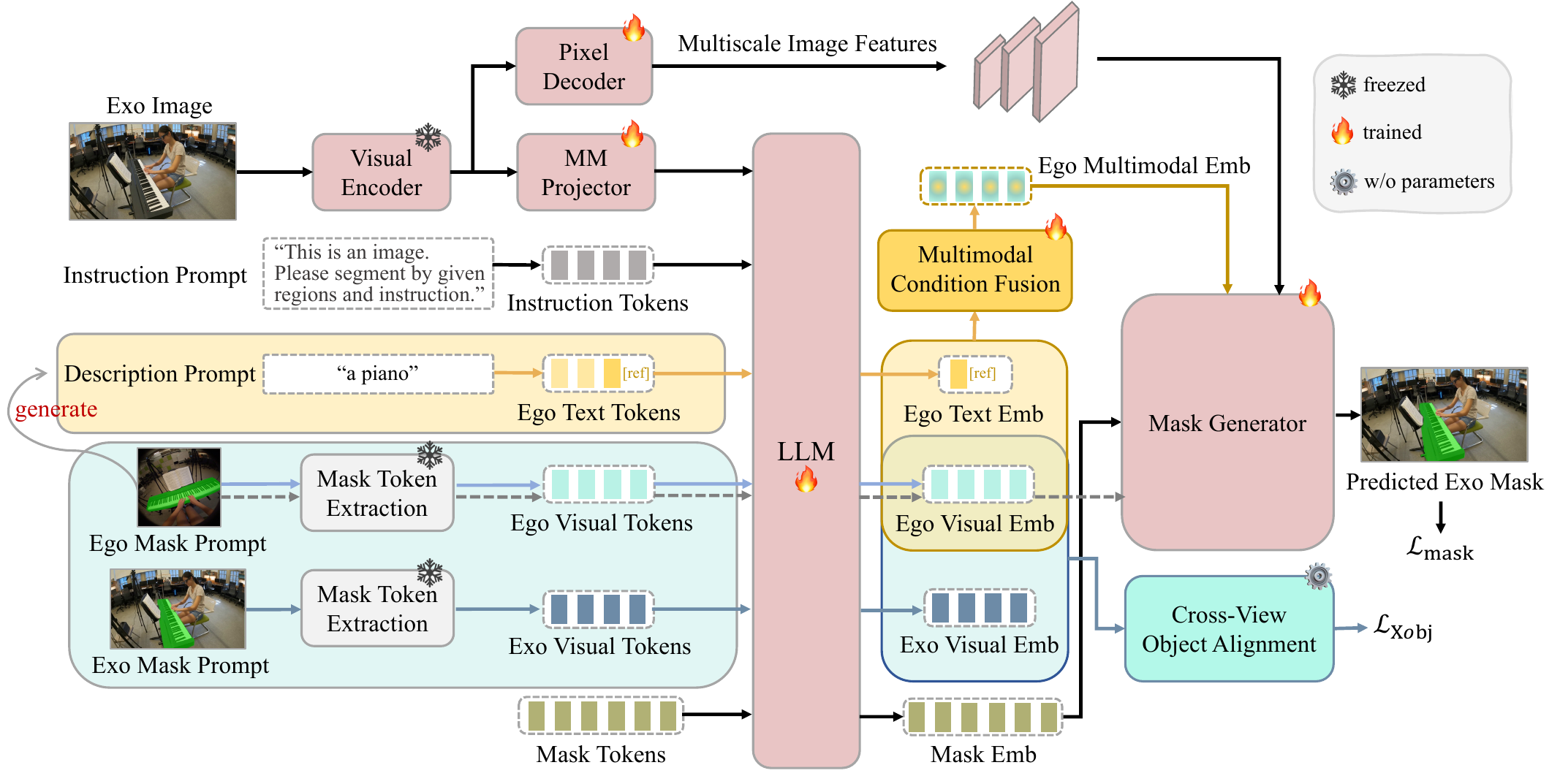}}
    \caption{Illustration of the proposed approach~\cite{fu2024objectrelator} for Ego-Exo4D Correspondence. Ego→Exo is shown as an example. We take PSALM as the baseline, with its modules indicated in pink, and integrate two novel modules (Multimodal Condition Fusion and Cross-View Object Alignment), which are highlighted with orange and green, respectively. 
    \label{fig:framework}}
    \vspace{-0.1in}
\end{figure}

As introduced in Sec.\ref{sec:intro}, we build our ObjectRelator method~\cite{fu2024objectrelator} upon the PSALM baseline. An overview of our framework is shown in Fig.~\ref{fig:framework}, where we mark the components inherited from PSALM in pink, including the Visual Encoder, Multimodal (MM) Projector, LLM, Pixel Decoder, Mask Generator, the mask loss function $\mathcal{L}_{mask}$, and the token extraction strategy. Our newly introduced components, Multimodal Condition Fusion (MCFuse) and Cross-View Object Alignment (XObjAlign), are highlighted in orange and green, respectively. These modules are specifically designed to address the challenges of the Ego-Exo4D Object Correspondence task, particularly in enhancing cross-view object localization and enforcing object-level consistency.

In particular, MCFuse improves the baseline by incorporating self-generated textual descriptions of the target objects and fusing both visual and textual cues to enhance segmentation localization. XObjAlign addresses the performance drop caused by the significant view shift between ego and exo perspectives by enforcing object-level consistency in the condition space through a self-supervised alignment strategy.

In the overall pipeline, illustrated using the Ego→Exo task as an example, the model takes a target (exo) video frame, along with an instruction prompt, object mask(s) derived from the query (ego) view, and mask tokens as inputs. The Visual Encoder, MM Projector, and Pixel Decoder are first applied to the target frame to extract both projected visual features and multiscale image features. The MCFuse module begins by generating a textual description prompt (e.g., "a piano") based on the query frame. Both the query mask prompt and the generated text prompt are then fed into the LLM to produce their respective embedding features, which are subsequently fused into the MCFuse to generate the multimodal embedding.
In parallel, the XObjAlign module introduces a self-supervised mechanism to improve cross-view consistency. It feeds the target object mask into the LLM to obtain a target visual embedding and enforces alignment between this and the query object embedding. This constraint helps the model learn more robust, view-invariant object representations.
Finally, the Mask Generator produces the segmentation mask prediction for the target view, using the multiscale image features from the target frame, the fused multimodal embedding, and the mask token embeddings as inputs.

\subsection{Proposed Modules}
\noindent\textbf{Multimodal Condition Fusion (MCFuse).} 
To achieve effective multimodal fusion of visual masks and textual descriptions, MCFuse addresses two core challenges: 1) How to automatically generate textual descriptions from the query frame, especially when the target object is small, blurry, or embedded in a cluttered scene. 2) How to effectively integrate the visual and textual modalities, which are inherently different in nature.

For the first challenge, following the task constraints, we do not use ground-truth textual annotations. Instead, we leverage state-of-the-art vision-language models such as LLaVA~\cite{liu2023llava}. However, directly applying LLaVA proves unreliable in complex cases. To address this, we design a tailored prompt strategy that combines the input image with the object mask, highlighting the target object while retaining the full scene context. This masked input helps guide LLaVA’s attention to the relevant region and significantly improves the quality and feasibility of the generated descriptions. For the second challenge, we empirically find that naive concatenation of visual and textual embeddings leads to suboptimal performance. We thus design a residual fusion architecture, where the visual modality serves as the primary branch, and the text modality contributes via a residual connection. A learnable fusion weight is introduced to automatically balance the contributions of each modality, avoiding the need for manual tuning. Together, these designs enable MCFuse to successfully incorporate textual cues and significantly enhance object localization.

\noindent\textbf{Cross-View Object Alignment (XObjAlign).} 
To ensure strong performance under large view shifts, our XObjAlign module is designed to enhance the model's robustness by enforcing consistency across different viewpoints. While the idea of using self-supervised learning for cross-view consistency is not new, we are the first to explore it in the context of the Ego-Exo4D Correspondence task. Specifically, we propose to apply this consistency at the object level, where it directly impacts segmentation results by improving the quality of the conditioning input. The key intuition is that if our model can successfully align the query and target object embeddings, it effectively eliminates the view-shift challenge, since this would be equivalent to using a condition derived from the target view itself to segment objects within that same view.

To implement this idea, we reuse the ground-truth object mask from the current target frame and feed it into the LLM to obtain the target object visual embedding. We then compute the cross-view object consistency loss $\mathcal{L}_{Xobj}$ by calculating the Euclidean distance between the visual embeddings of the query and target objects.

For detailed architecture and configurations of the modules, please refer to our prior work~\cite{fu2024objectrelator}.

\subsection{Training and Inference.}
The pretrained weights for the baseline modules are initialized from PSALM. To ensure effective optimization of the newly introduced MCFuse module, we first train it using the segmentation loss $\mathcal{L}_{mask}$. We then jointly train all modules (indicated by the fire icons in Fig.~\ref{fig:framework}), except the Visual Encoder, using the combined loss $\mathcal{L}_{mask} + \mathcal{L}_{Xobj}$. It is worth noting that XObjAlign does not introduce any additional parameters and is only applied during the training phase.

\section{Experiments}
\label{sec:Experiments}

\subsection{Datasets and Experimental Setup}
\noindent\textbf{Datasets.} 
Following the official challenge protocol, we evaluate our proposed method on the Ego-Exo4D dataset~\cite{grauman2024ego}. Ego-Exo4D consists of 1,335 video takes and over 1.8 million annotated object masks, covering diverse domains such as Cooking, Bike Repair, Healthcare, Music, Basketball, and Soccer. We follow the standard train/validation/test splits provided by the challenge, with final evaluation conducted on the test set.

\noindent\textbf{Experimental Setup.} 
We use a single model trained jointly on both the Ego→Exo and Exo→Ego training sets to evaluate performance on both corresponding test sets. In the first training stage, we use only 1/20 of the joint training data to train the MCFuse module. In the second stage, we train all components (excluding the Visual Encoder) using the full joint training set. Each stage runs for 3 epochs with a batch size of 12, using the AdamW optimizer. Training is performed on 4 NVIDIA A100 GPUs.

\subsection{Comparison Results}
The comparison results are summarized in Tab.~\ref{tab:main}. Note that XSegTx is the official baseline provided by the challenge organizers, while the other entries represent submissions from participating teams. For both the Ego→Exo and Exo→Ego tasks, we report IoU, VA, LE, and CA metrics.

Our proposed method clearly outperforms the XSegTx baseline, improving IoU from 0.19 to 0.35 on Ego→Exo and from 0.27 to 0.40 on Exo→Ego, and ranks second overall on the leaderboard. These results strongly validate the effectiveness of our approach. Moreover, in terms of Visibility Accuracy (VA), our method significantly surpasses all other competitors, achieving 96\% on Ego→Exo and 97\% on Exo→Ego, respectively.

\begin{table*}[h]
\centering 
\resizebox{0.95\textwidth}{!}{
\begin{tabular}{llcccc|llcccc}
\toprule
\multicolumn{6}{c|}{\textbf{Ego→Exo (Ego as Query)}} & \multicolumn{6}{c}{\textbf{Exo→Ego (Exo as Query)}}
 \\ \midrule[0.6pt]
\textbf{Method} & \textbf{Team} & \textbf{IoU}$\uparrow$ & \textbf{VA}$\uparrow$ & \textbf{LE}$\downarrow$ & \textbf{CA}$\uparrow$  & \textbf{Method} &  \textbf{Team} & \textbf{IoU}$\uparrow$ & \textbf{VA}$\uparrow$  & \textbf{LE}$\downarrow$ & \textbf{CA}$\uparrow$  \\ \midrule[0.6pt] 
O-MAMA & OMama\_bis & \cellcolor{pink!40}\textbf{0.43}& 0.50	& \cellcolor{pink!40}\textbf {0.03}	& \cellcolor{pink!40}\textbf{0.59}  & O-MAMA & OMama\_bis &	\cellcolor{pink!40}\textbf{0.44}& 0.50	& 0.08& 0.52 \\
\textbf{ObjectRelator (Ours)} & Winning Bunnies & 	0.35& \cellcolor{pink!40}\textbf{0.96}	& 0.04	& 0.54& \textbf{ObjectRelator (Ours)} & Winning Bunnies &  0.40	 & \cellcolor{pink!40}\textbf{0.97}	 & \cellcolor{pink!40}\textbf{0.07}	& 0.50	\\
- & Touching\_fish & 0.32& 0.68 & 	0.04	& 0.57  & - & Touching\_fish & 	0.41 & 	0.69& 0.09	& \cellcolor{pink!40}\textbf{0.53} \\
XSegTx & Host & 0.19 & 0.66 & 	0.07& 0.39	 & XSegTx & Host &0.27	& 0.82	& 0.10 & 0.36\\  
\bottomrule[0.95pt]
\end{tabular}
}
\caption{Comparison results on Ego-Exo4D test sets. Results are from the Ego-Exo4D correspondence challenge leaderboard.}
\label{tab:main}
\end{table*}

\begin{figure}[h]
    \centering
           \vspace{-0.1in}
       {\includegraphics[width=1.\linewidth]{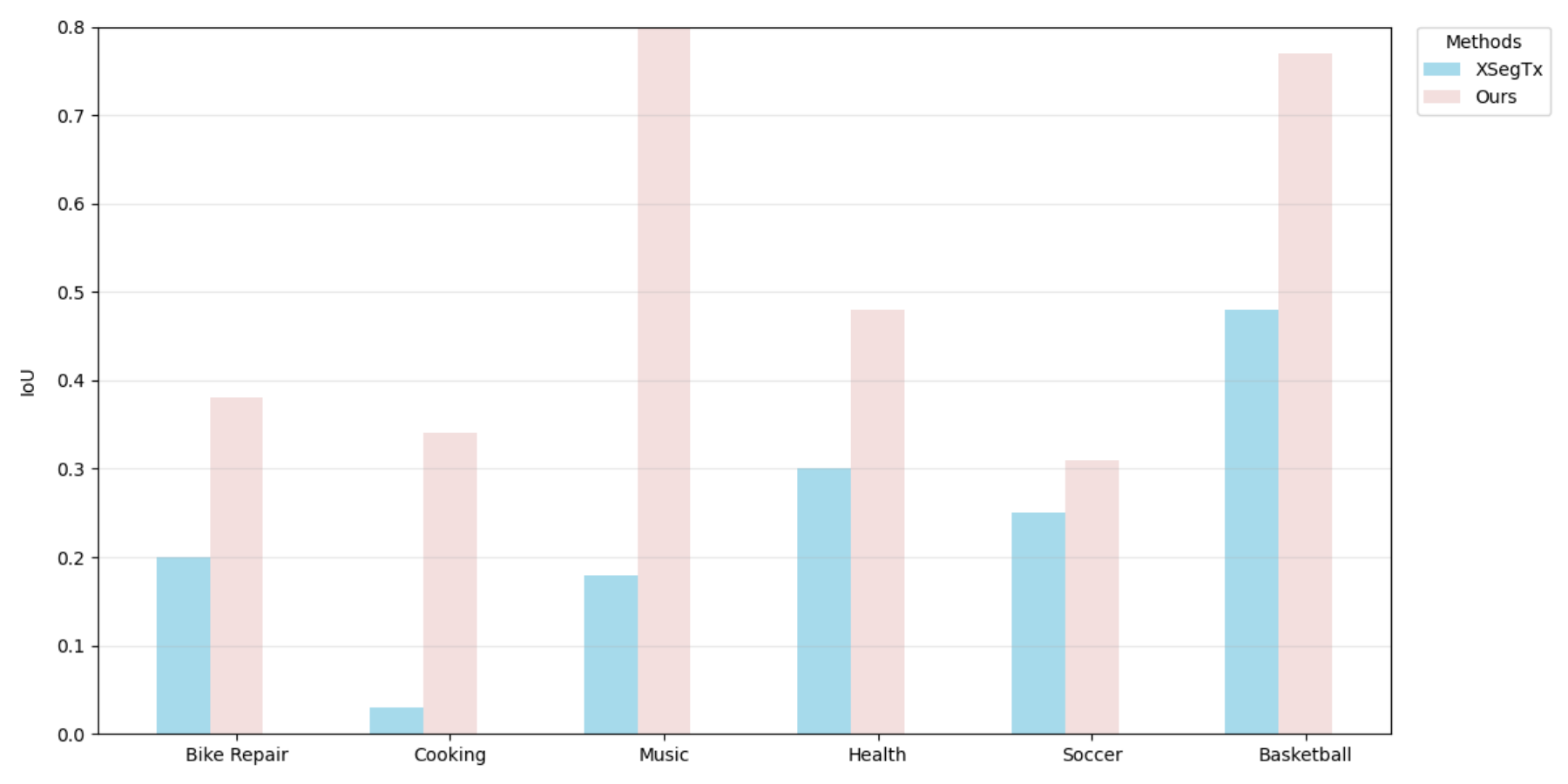}}
       \vspace{-0.2in}
    \caption{IoU of our method on different scenarios, Ego→Exo.
    \label{fig:analysis}}
    \vspace{-0.1in}
\end{figure}

\noindent\textbf{Performance Analysis Across Different Scenarios.} 
Since the Ego-Exo4D benchmark spans six diverse scenarios, we analyze performance (primarily IoU) across each scenario individually. The results—using the Ego→Exo setting as an example, are summarized in Fig.~\ref{fig:analysis}.

From the results, we observe the following: 1) Across all scenarios, our proposed method consistently and significantly outperforms the XSegTx baseline. 2) Performance varies by scenario, with higher IoU scores observed in domains such as Music and Basketball, likely due to the presence of relatively simple and easily distinguishable objects (e.g., sheet music, basketballs). In contrast, scenarios like Cooking present more complex and cluttered environments, making object correspondence more challenging.

\subsection{Ablation Studies}
To validate the effectiveness of our proposed modules, MCFuse and XObjAlign, we conduct an ablation study by incrementally adding each module to the baseline. Results for the Ego→Exo setting are reported in Tab.~\ref{tab:abla-modules}.

The results show that both modules contribute significantly to performance improvement: MCFuse increases the baseline IoU by 3\%, and XObjAlign adds an additional 4\% improvement. When combined, our full model achieves the best performance, boosting IoU from 30\% to 35\%, which further validates the effectiveness of our proposed approach.

\begin{table}[h]
\centering 
\resizebox{0.4\textwidth}{!}{
\begin{tabular}{l|ccc}
\toprule
\textbf{Method} & \textbf{MCFuse} & \textbf{XObjAlign} &  \textbf{Ego→Exo} $\uparrow$ \\ \hline
Base & \ding{55}  &  \ding{55}  &  0.30  \\  \hline
+MCFuse   &  \ding{52} &  \ding{55}  & 0.33  \\  \hline
+XObjAlign&  \ding{55} &  \ding{52}  &  0.34 \\ \hline
\cellcolor{pink!40}\textbf{ObjectRelator}   &  \cellcolor{pink!40}\ding{52}  &  \cellcolor{pink!40}\ding{52}  &  \cellcolor{pink!40}\textbf{0.35}   \\
\bottomrule
\end{tabular}
}
\caption{Effectiveness of proposed modules (IoUs on Ego→Exo).}
\label{tab:abla-modules}
\end{table}

\subsection{Visualization Results}
\begin{figure}[t]
    \centering
       {\includegraphics[width=1.\linewidth]{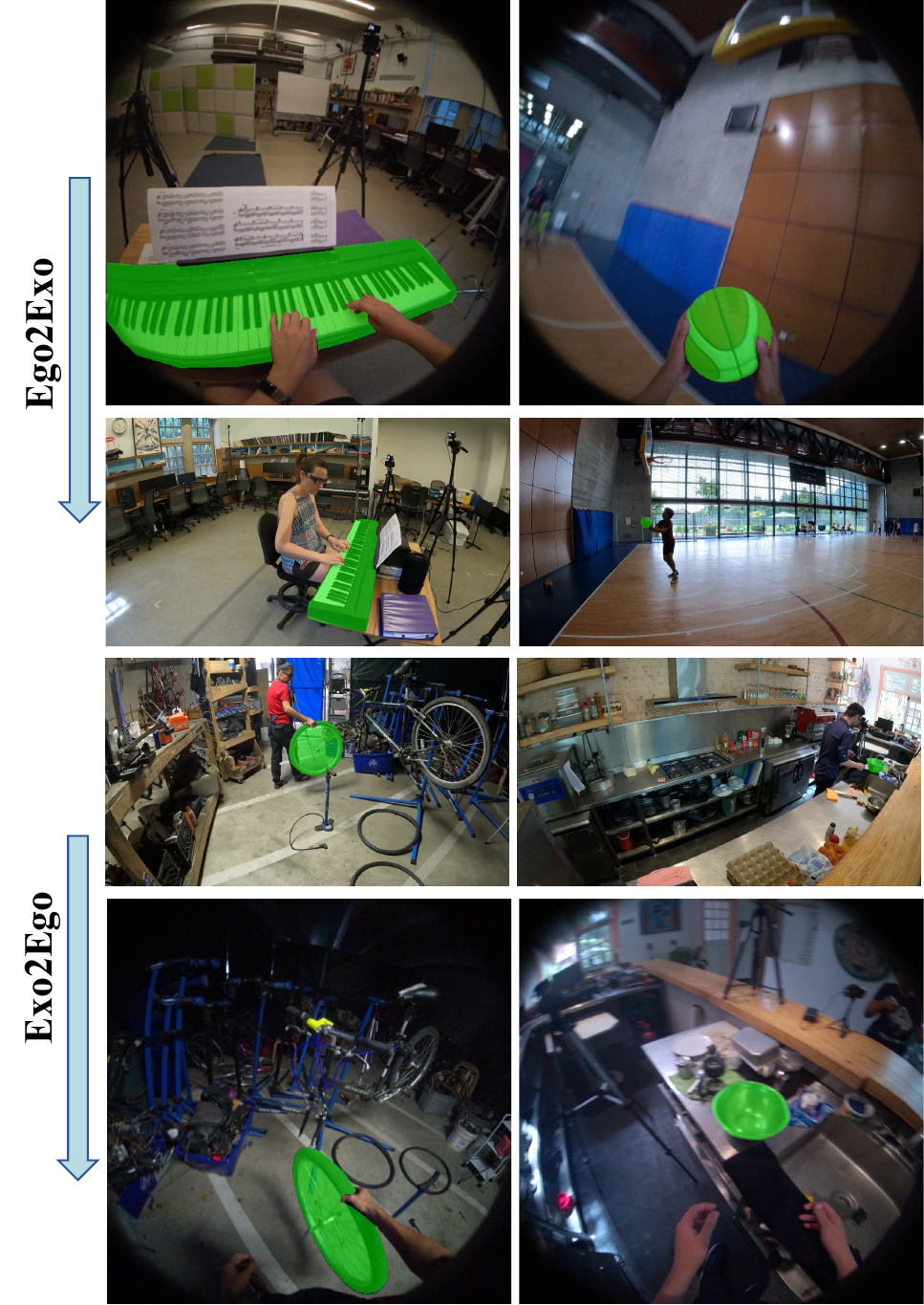}}
       \vspace{-0.1in}
    \caption{Visualization of our method on testsets.
    \label{fig:vis}}
    \vspace{-0.1in}
\end{figure}

In Fig.~\ref{fig:vis}, we visualize predictions from our method on both Ego→Exo and Exo→Ego tasks, with examples randomly sampled from the test set. The results demonstrate that, given a query view, our method can accurately predict the corresponding object in the target view.

\section{Conclusion and Limitations}
\label{sec:conclusion}
In this technical report, we presented ObjectRelator, a cross-view segmentation framework for the Ego-Exo Object Correspondence task, integrating two novel modules: MCFuse for visual-text condition fusion and XObjAlign for cross-view consistent alignment. Our method achieves second place in IoU and first in VA on the leaderboard, demonstrating strong performance across diverse scenarios.

\noindent\textbf{Limitations.} The model may produce incomplete masks in cluttered scenes and does not yet fully exploit the temporal information available in the videos.

\section{Acknowledgments}
This research was partially funded by the Ministry of Education and Science of Bulgaria (support for INSAIT, part of the Bulgarian National Roadmap for Research Infrastructure). This project was supported with computational resources provided by Google Cloud Platform (GCP).

{
    \small
    \bibliographystyle{ieeenat_fullname}
    \bibliography{main}

\begin{thebibliography}{21}
\providecommand{\natexlab}[1]{#1}
\providecommand{\url}[1]{\texttt{#1}}
\expandafter\ifx\csname urlstyle\endcsname\relax
  \providecommand{\doi}[1]{doi: #1}\else
  \providecommand{\doi}{doi: \begingroup \urlstyle{rm}\Url}\fi

\bibitem[An et~al.(2023)An, Sun, Wu, Tang, and Van~Gool]{an2023temporal}
Zhaochong An, Guolei Sun, Zongwei Wu, Hao Tang, and Luc Van~Gool.
\newblock Temporal-aware hierarchical mask classification for video semantic segmentation.
\newblock \emph{arXiv preprint arXiv:2309.08020}, 2023.

\bibitem[Br{\"o}dermann et~al.(2025)Br{\"o}dermann, Sakaridis, Fu, and Van~Gool]{brodermann2025cafuser}
Tim Br{\"o}dermann, Christos Sakaridis, Yuqian Fu, and Luc Van~Gool.
\newblock Cafuser: Condition-aware multimodal fusion for robust semantic perception of driving scenes.
\newblock \emph{IEEE Robotics and Automation Letters}, 2025.

\bibitem[Cheng et~al.(2022)Cheng, Misra, Schwing, Kirillov, and Girdhar]{cheng2022masked}
Bowen Cheng, Ishan Misra, Alexander~G Schwing, Alexander Kirillov, and Rohit Girdhar.
\newblock Masked-attention mask transformer for universal image segmentation.
\newblock In \emph{Proceedings of the IEEE/CVF conference on computer vision and pattern recognition}, pages 1290--1299, 2022.

\bibitem[Fu et~al.(2024)Fu, Wang, Fu, Paudel, Huang, and Van~Gool]{fu2024objectrelator}
Yuqian Fu, Runze Wang, Yanwei Fu, Danda~Pani Paudel, Xuanjing Huang, and Luc Van~Gool.
\newblock Objectrelator: Enabling cross-view object relation understanding in ego-centric and exo-centric videos.
\newblock \emph{arXiv preprint arXiv:2411.19083}, 2024.

\bibitem[Grauman et~al.(2024)Grauman, Westbury, Torresani, Kitani, Malik, Afouras, Ashutosh, Baiyya, Bansal, Boote, et~al.]{grauman2024ego}
Kristen Grauman, Andrew Westbury, Lorenzo Torresani, Kris Kitani, Jitendra Malik, Triantafyllos Afouras, Kumar Ashutosh, Vijay Baiyya, Siddhant Bansal, Bikram Boote, et~al.
\newblock Ego-exo4d: Understanding skilled human activity from first-and third-person perspectives.
\newblock In \emph{Proceedings of the IEEE/CVF Conference on Computer Vision and Pattern Recognition}, pages 19383--19400, 2024.

\bibitem[Guo et~al.(2024)Guo, Li, Huang, Hong, Zhou, Chen, Li, Jiang, Zhang, and Zhang]{guo2024x}
Pinxue Guo, Wanyun Li, Hao Huang, Lingyi Hong, Xinyu Zhou, Zhaoyu Chen, Jinglun Li, Kaixun Jiang, Wei Zhang, and Wenqiang Zhang.
\newblock X-prompt: Multi-modal visual prompt for video object segmentation.
\newblock In \emph{Proceedings of the 32nd ACM International Conference on Multimedia}, pages 5151--5160, 2024.

\bibitem[He et~al.(2017)He, Gkioxari, Doll{\'a}r, and Girshick]{he2017mask}
Kaiming He, Georgia Gkioxari, Piotr Doll{\'a}r, and Ross Girshick.
\newblock Mask r-cnn.
\newblock In \emph{Proceedings of the IEEE international conference on computer vision}, pages 2961--2969, 2017.

\bibitem[Kirillov et~al.(2023)Kirillov, Mintun, Ravi, Mao, Rolland, Gustafson, Xiao, Whitehead, Berg, Lo, et~al.]{kirillov2023segment}
Alexander Kirillov, Eric Mintun, Nikhila Ravi, Hanzi Mao, Chloe Rolland, Laura Gustafson, Tete Xiao, Spencer Whitehead, Alexander~C Berg, Wan-Yen Lo, et~al.
\newblock Segment anything.
\newblock In \emph{Proceedings of the IEEE/CVF International Conference on Computer Vision}, pages 4015--4026, 2023.

\bibitem[Lai et~al.(2024)Lai, Tian, Chen, Li, Yuan, Liu, and Jia]{lai2024lisa}
Xin Lai, Zhuotao Tian, Yukang Chen, Yanwei Li, Yuhui Yuan, Shu Liu, and Jiaya Jia.
\newblock Lisa: Reasoning segmentation via large language model.
\newblock In \emph{Proceedings of the IEEE/CVF Conference on Computer Vision and Pattern Recognition}, pages 9579--9589, 2024.

\bibitem[Li et~al.(2024{\natexlab{a}})Li, Guo, Zhou, Hong, He, Zheng, Zhang, and Zhang]{li2024onevos}
Wanyun Li, Pinxue Guo, Xinyu Zhou, Lingyi Hong, Yangji He, Xiangyu Zheng, Wei Zhang, and Wenqiang Zhang.
\newblock Onevos: unifying video object segmentation with all-in-one transformer framework.
\newblock In \emph{European Conference on Computer Vision}, pages 20--40. Springer, 2024{\natexlab{a}}.

\bibitem[Li et~al.(2024{\natexlab{b}})Li, Yuan, Li, Ding, Wu, Zhang, Li, Chen, and Loy]{li2024omg}
Xiangtai Li, Haobo Yuan, Wei Li, Henghui Ding, Size Wu, Wenwei Zhang, Yining Li, Kai Chen, and Chen~Change Loy.
\newblock Omg-seg: Is one model good enough for all segmentation?
\newblock In \emph{Proceedings of the IEEE/CVF Conference on Computer Vision and Pattern Recognition}, pages 27948--27959, 2024{\natexlab{b}}.

\bibitem[Liu et~al.(2023)Liu, Li, Wu, and Lee]{liu2023llava}
Haotian Liu, Chunyuan Li, Qingyang Wu, and Yong~Jae Lee.
\newblock Visual instruction tuning, 2023.

\bibitem[Rasheed et~al.(2024)Rasheed, Maaz, Shaji, Shaker, Khan, Cholakkal, Anwer, Xing, Yang, and Khan]{rasheed2024glamm}
Hanoona Rasheed, Muhammad Maaz, Sahal Shaji, Abdelrahman Shaker, Salman Khan, Hisham Cholakkal, Rao~M Anwer, Eric Xing, Ming-Hsuan Yang, and Fahad~S Khan.
\newblock Glamm: Pixel grounding large multimodal model.
\newblock In \emph{Proceedings of the IEEE/CVF Conference on Computer Vision and Pattern Recognition}, pages 13009--13018, 2024.

\bibitem[Ren et~al.(2024)Ren, Huang, Wei, Zhao, Fu, Feng, and Jin]{ren2024pixellm}
Zhongwei Ren, Zhicheng Huang, Yunchao Wei, Yao Zhao, Dongmei Fu, Jiashi Feng, and Xiaojie Jin.
\newblock Pixellm: Pixel reasoning with large multimodal model.
\newblock In \emph{Proceedings of the IEEE/CVF Conference on Computer Vision and Pattern Recognition}, pages 26374--26383, 2024.

\bibitem[Wu et~al.(2023)Wu, Wang, Zhou, An, Jiang, Demonceaux, Sun, and Timofte]{wu2023object}
Zongwei Wu, Jingjing Wang, Zhuyun Zhou, Zhaochong An, Qiuping Jiang, C{\'e}dric Demonceaux, Guolei Sun, and Radu Timofte.
\newblock Object segmentation by mining cross-modal semantics.
\newblock In \emph{Proceedings of the 31st ACM International Conference on Multimedia}, pages 3455--3464, 2023.

\bibitem[Yan et~al.(2023)Yan, Jiang, Wu, Wang, Luo, Yuan, and Lu]{yan2023universal}
Bin Yan, Yi Jiang, Jiannan Wu, Dong Wang, Ping Luo, Zehuan Yuan, and Huchuan Lu.
\newblock Universal instance perception as object discovery and retrieval.
\newblock In \emph{Proceedings of the IEEE/CVF Conference on Computer Vision and Pattern Recognition}, pages 15325--15336, 2023.

\bibitem[Zhang et~al.(2024)Zhang, Ma, Zhang, and Bai]{zhang2024psalm}
Zheng Zhang, Yeyao Ma, Enming Zhang, and Xiang Bai.
\newblock Psalm: Pixelwise segmentation with large multi-modal model.
\newblock In \emph{European Conference on Computer Vision}, pages 74--91. Springer, 2024.

\bibitem[Zheng et~al.(2024)Zheng, Lyu, and Wang]{zheng2024learning}
Xu Zheng, Yuanhuiyi Lyu, and Lin Wang.
\newblock Learning modality-agnostic representation for semantic segmentation from any modalities.
\newblock In \emph{European Conference on Computer Vision}, pages 146--165. Springer, 2024.

\bibitem[Zheng et~al.(2025)Zheng, Luo, Zhou, and Wang]{zheng2025distilling}
Xu Zheng, Yunhao Luo, Pengyuan Zhou, and Lin Wang.
\newblock Distilling efficient vision transformers from cnns for semantic segmentation.
\newblock \emph{Pattern Recognition}, 158:\penalty0 111029, 2025.

\bibitem[Zhou et~al.(2025)Zhou, Sun, Li, Fu, Benini, and Konukoglu]{zhou2025camsam2}
Yuli Zhou, Guolei Sun, Yawei Li, Yuqian Fu, Luca Benini, and Ender Konukoglu.
\newblock Camsam2: Segment anything accurately in camouflaged videos.
\newblock \emph{arXiv preprint arXiv:2503.19730}, 2025.

\bibitem[Zou et~al.(2024)Zou, Yang, Zhang, Li, Li, Wang, Wang, Gao, and Lee]{zou2024segment}
Xueyan Zou, Jianwei Yang, Hao Zhang, Feng Li, Linjie Li, Jianfeng Wang, Lijuan Wang, Jianfeng Gao, and Yong~Jae Lee.
\newblock Segment everything everywhere all at once.
\newblock \emph{Advances in Neural Information Processing Systems}, 36, 2024.

\end{thebibliography}
}

\end{document}